\documentclass[letterpaper, 10 pt, conference]{ieeeconf}  

\usepackage{cite}
\usepackage{graphicx}
\usepackage{booktabs}
\usepackage{amssymb}
\usepackage{amsmath}
\usepackage{hyperref}
\usepackage{floatrow}
\newfloatcommand{capbtabbox}{table}[\captop][\FBwidth]

\newcommand{\citet}[1]{\cite{#1}}

\IEEEoverridecommandlockouts                              

\title{\LARGE \bf
In-Hand Object Pose Estimation via Visual-Tactile Fusion
}

\author{Felix Nonnengießer$^{1}$, Alap Kshirsagar$^{2}$, Boris Belousov$^{3}$ and Jan Peters$^{2,3,4,5}$
\thanks{\textsuperscript{1}Department of Computer Science, Goethe Universität Frankfurt, Germany. {\tt felixnon@cs.uni-frankfurt.de} }
\thanks{\textsuperscript{2}Intelligent Autonomous Systems Lab, Department of Computer Science, TU Darmstadt, Germany }
\thanks{\textsuperscript{3}German Research Center for AI (DFKI)}
\thanks{\textsuperscript{4}Centre for Cognitive Science, TU Darmstadt}
\thanks{\textsuperscript{5}Hessian Center for Artificial Intelligence (Hessian.AI), Darmstadt}
\thanks{We thank Hessisches Ministerium für Wissenschaft \& Kunst for the DFKI grant and ``The Adaptive Mind'' grant.}
}

\begin{document}

\maketitle
\thispagestyle{empty}
\pagestyle{empty}

\begin{abstract}
Accurate in-hand pose estimation is crucial for robotic object manipulation, but visual occlusion remains a major challenge for vision-based approaches. This paper presents an approach to robotic in-hand object pose estimation, combining visual and tactile information to accurately determine the position and orientation of objects grasped by a robotic hand. We address the challenge of visual occlusion by fusing visual information from a wrist-mounted RGB-D camera with tactile information from vision-based tactile sensors mounted on the fingertips of a robotic gripper. Our approach employs a weighting and sensor fusion module to combine point clouds from heterogeneous sensor types and control each modality's contribution to the pose estimation process. We use an augmented Iterative Closest Point (ICP) algorithm adapted for weighted point clouds to estimate the 6D object pose. Our experiments show that incorporating tactile information significantly improves pose estimation accuracy, particularly when occlusion is high. Our method achieves an average pose estimation error of 7.5 mm and 16.7 degrees, outperforming vision-only baselines by up to 20\%. We also demonstrate the ability of our method to perform precise object manipulation in a real-world insertion task. 
\end{abstract}

\section{Introduction}
\label{chap:intro}

In-hand pose estimation describes the process of determining the position and orientation of an object held within a robotic hand.  
Visuotactile-based in-hand pose estimation methods employ one or more cameras along with tactile sensors in order to determine the pose of a grasped object.

A precise pose estimation for objects held by a robotic hand is essential for improving the accuracy and efficiency of robotic manipulation and assembly tasks. This ability is particularly relevant in industrial and manufacturing applications where robots need to accurately grasp, manipulate and assemble objects. Especially in uncontrolled environments, the success of such tasks is often determined by the accuracy of pose estimation~\cite{litvak2019learning, quentin2023industrial}.

\begin{figure}[ht]
     \centering
     \includegraphics[width=\linewidth]{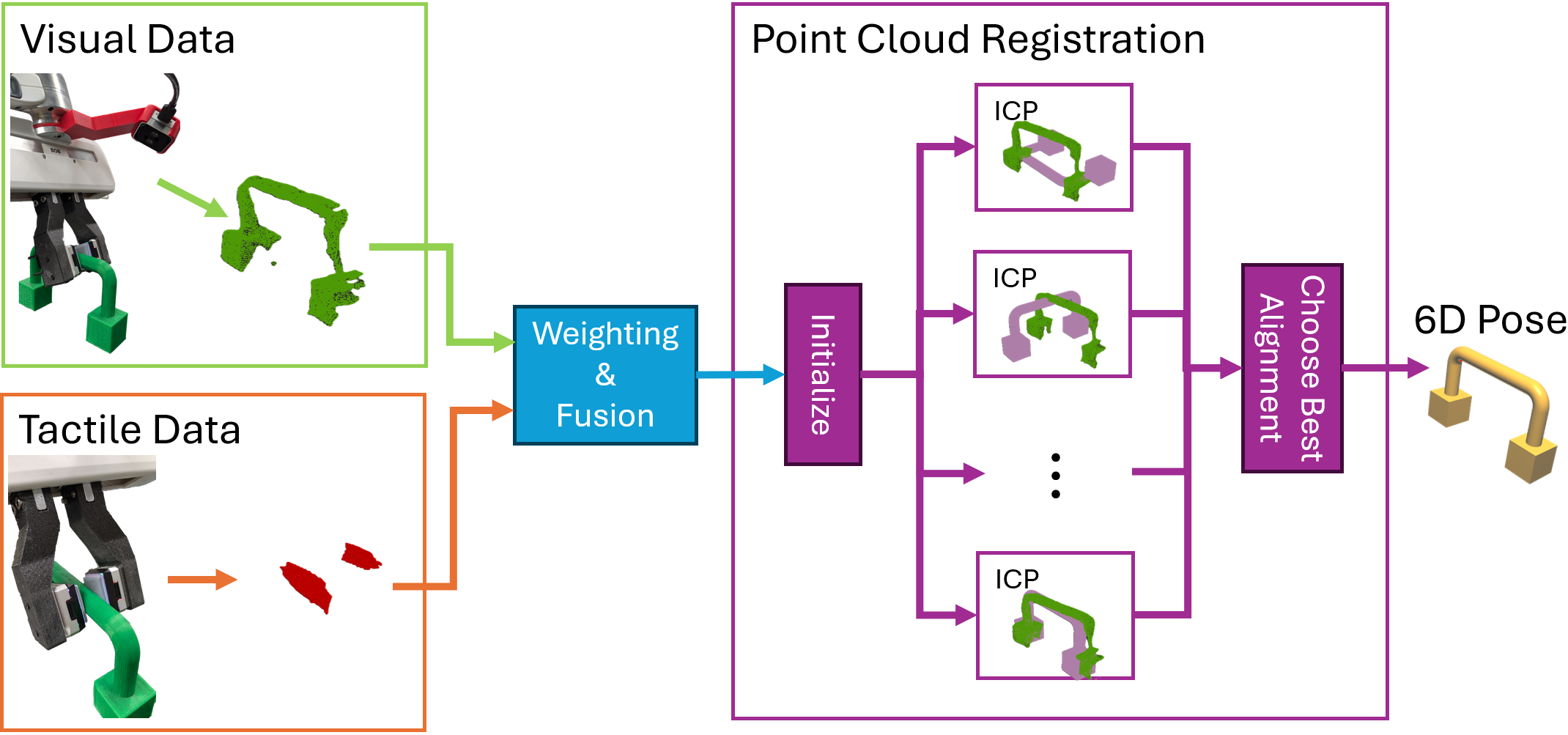}
     \caption{Overview of the proposed multimodal pose estimation framework. The visual data (green) and tactile data (orange) are separately preprocessed before the point clouds are combined and weighted (blue). The Registration module (purple) performs an ICP algorithm using multiple initializations to determine an estimate of the 6D object pose. }
     \label{fig:pipeline}
\end{figure}

Though cameras are often used for object pose estimation, visual occlusion of the object can affect the success and accuracy of the estimation. The object may be occluded by other items in the environment, the robot itself, or particularly by the robotic gripper or hand during grasping. In such scenarios of heavy occlusion, the performance of pose estimation approaches relying only on visual information often diminishes~\cite{billard2019trends, wen2020robust}.
This limitation spurred interest in supplementing the visual information with tactile sensing. 

Tactile sensors mounted to the robotic hand are in direct physical contact with grasped objects and are therefore not affected by occlusion. They provide precise information about the local geometry of the object even when visual information is sparse due to occlusion or difficult lighting conditions. However, tactile sensors have a relatively small field of view, limited by the contact area between the sensor and the object.

Existing work in visuotactile-based approaches for in-hand pose estimation has shown promising results in combining the strengths of visual and tactile sensing~\cite{izatt2017tracking,liu2023enhancing}. Researchers have proposed various methods to estimate object poses from visual and tactile data, including the design of extended Kalman filters~\cite{izatt2017tracking}, training of neural networks~\cite{anzai2020deep,dikhale2022visuotactile}, and comparing encoded sensor data against a dense set of simulated object poses and their respective sensor readings~\cite{gao2023hand}.
As in our work, many of these methods use vision-based tactile sensors like GelSight~\cite{yuan2017gelsight} and DIGIT~\cite{lambeta2020digit}, as they provide high-resolution information on local geometry.

Many current pose estimation approaches rely on deep learning techniques or auto-encoders, necessitating extensive training and large datasets. However, works that do not employ learning-based algorithms typically focus on in-hand object tracking and require a prior estimate of the object pose. To address this gap, we develop a method for one-shot pose estimation that does not rely on learning-based algorithms or initial pose estimates. By utilizing a registration algorithm, as shown in Fig.~\ref{fig:pipeline}, we aim to create a flexible and adaptable approach for in-hand pose estimation that can be applied to new objects without extensive training or data collection.

Our contributions are:
\begin{enumerate}
    \item We propose a visuo-tactile in-hand pose estimation method that does not require extensive (re-)training, making it easily adaptable to new objects.
    \item We implement and investigate a weighted multimodal point cloud registration approach that weights and combines visual and tactile sensor data to balance the contribution of each modality.
    \item We analyze the impact and demonstrate the benefits of tactile sensing for in-hand pose estimation particularly in cases of significant visual occlusion.
\end{enumerate}

\section{Related Work}
\label{chap:relatedwork}

This section reviews various approaches in robotic in-hand pose estimation, categorizing them by the employed sensor modalities: vision-based, tactile-based, and visuotactile-based methods. 

\subsection{Vision-Based Approaches}

Many vision-based pose estimation approaches focus on objects in isolation or cluttered environments, instead of specifically addressing the challenges of in-hand pose estimation, where the object is held by a robotic gripper or hand. Such general 6D object pose estimation approaches typically employ RGB or RGB-D cameras and a variety of methods, including feature and template matching \cite{li2018deepim, zakharov2019dpod}, point cloud registration \cite{besl1992method}, and machine learning approaches \cite{xiang2017posecnn, tekin2018real}.
However, in-hand object pose estimation presents additional challenges, primarily due to significant occlusion caused by the grasping hand. 

Recent research addresses these challenges through different approaches. Instead of tracking the object pose directly, Raessa et al.~\cite{raessa2019visually} track AR (Augmented Reality) markers attached to objects to estimate their pose during manipulation. While AR markers can generally be localized with high precision, they need to remain within the camera's field of view during the manipulation and it might not be feasible to attach markers to objects in every scenario. Wen et al.~\cite{wen2020robust} proposed an occlusion-aware pose estimation method for objects grasped by adaptive hands. By estimating the configuration of the robotic hand, they can isolate points of the object between the fingers from a point cloud obtained by an RGB-D camera. The object segment is used to generate pose hypotheses that are subsequently filtered based on physical reasoning. Liu et al.~\cite{liu2021robust} track the pose of a pen held by a robotic gripper when drawing on uneven surfaces. They use a conventional Iterative Closest Point (ICP) algorithm to estimate the pen's pose, based on a point cloud obtained by a depth camera. However, they improve the precision of the point cloud matching step by unevenly sampling the object mesh dependent on the camera's viewpoint. Pfanne et al.~\cite{pfanne2018fusing} use visual features like edges and corners observed by a monocular camera in combination with state information of the manipulator like finger positions and torque measurements. This information is fused using an extended Kalman filter (EKF) to determine the grasped object's pose.

Vision-based methods can effectively estimate the pose of an object, but their performance can be significantly compromised by occlusion when objects are grasped by a robotic hand. To address this limitation, we combine visual data with precise, high-resolution tactile information obtained from vision-based tactile sensors. These sensors are in direct contact with the object and thus are not affected by occlusion.

\subsection{Tactile-Based Approaches}

Various kinds of tactile sensors are employed in tactile-based object pose estimation methods. Galaiya et al~\cite{galaiya2023exploring} use multiple compliant tactile sensors attached to the robotic fingers that detect pressure and vibration caused by a grasped object. The authors of~\cite{driels1986pose} introduced an array of contact sensors that can be attached to a robotic manipulator to infer an object's pose. Lach et al.~\cite{lach2023placing} use piezo-resisitve tactile sensors to correct the orientation of a grasped object for stable placement. However, new developments in camera-based high-resolution tactile sensing technologies such as GelSight~\cite{yuan2017gelsight} and DIGIT~\cite{lambeta2020digit} allowed for various new approaches in the domain of tactile-based object pose estimation.

Li et al.~\cite{li2014localization} were able to accurately localize and manipulate small objects employing tactile sensors only.
They used GelSight sensors to create high-resolution feature maps of the surface of small objects such as coins or USB connectors. 
These prerecorded height maps can then be used to localize the object in a parallel gripper equipped with a GelSight sensor using a feature-based registration algorithm.

Instead of creating a tactile map of the entire object surface, Tac2Pose by Bauza et al.~\cite{bauza2023tac2pose} uses an extensive set of simulated contact shapes that an object could potentially produce when in contact with a tactile sensor. 
These contact shapes are precomputed for a dense set of different poses which can then be matched in an embedded space with the actual sensor reading to estimate a probability distribution over possible object poses. 
While accurately localizing smaller parts with unique features, their results also show that the pose of larger objects with ambiguous features can be more difficult to estimate.

Caddeo et al.~\cite{caddeo2023collision} use multiple tactile DIGIT sensors to estimate the pose of several objects from the YCB objects dataset~\cite{berk2015ycb}. 
These objects are larger in size and provide generally less distinctive features than the objects used in the previously mentioned works.
In a preliminary step, they build a database of potential sensor positions by collecting a large number of simulated tactile images from the surface of the object.
During pose estimation, the actual sensor images are encoded and compared with the images in the database to find suitable candidates. 
The selection of candidate sensor positions is narrowed down using gradient descent optimization and a final pose is selected by ranking the remaining candidates based on whether or not they are in collision with the sensors.

Although methods employing vision-based tactile sensors have shown great success in estimating the pose of small objects with rich features, larger objects and those with ambiguous surface features remain challenging due to the sensors' limited field of view. We address this challenge by additionally employing an RGB-D camera, which provides a broader field of view.

\subsection{Visuotactile-Based Approaches}

Visuotactile-based approaches combine the strengths of both visual and tactile methods. While visual data provides rich global information, tactile sensors can offer precise local contact data. A challenge of visuotactile approaches is the fusion of these dissimilar sensor modalities.  Extended Kalman filters~\cite{izatt2017tracking} and neural network approaches~\cite{anzai2020deep, dikhale2022visuotactile} are commonly used to combine the sensor information. Other approaches obtain an initial pose estimation based on visual data to then refine it using the additional tactile information~\cite{liu2023enhancing, chaudhury2022using}.

Izatt et al.~\cite{izatt2017tracking} use an RGB-D camera in addition to a GelSight sensor mounted to the finger of a parallel gripper. 
The point clouds obtained from both sensors serve as input for a modified extended Kalman filter. 
The estimation of the pose is constrained by a set of rules to avoid poses that collide with the robot or poses where the object would be in areas that are known to be clear. 
The authors show that the addition of tactile data can significantly improve the tracking accuracy and that positional errors added to the camera data can be compensated for.

To determine the contribution of multiple sensor modalities, particularly an RGB camera and a GelSight tactile sensor, Anzai et al.~\cite{anzai2020deep} proposed a network architecture for estimating object pose changes that dynamically determines the reliability of each sensor module. 
The network, consisting of a unit for feature extraction, a deep gated multi-modal unit that determines each modality's contribution, and a final unit for estimating an object's pose change can be trained end-to-end and does not require a known 3D model when working with unseen objects. 
By dynamically adapting the contribution of individual sensor modules, the network can react to changing situations such as sensor failures and varying degrees of noise.

Gao et al.~\cite{gao2023hand} fuse multi-modal sensor information by separately encoding segmented camera images from two wrist-mounted RGB cameras and the tactile images from two finger-mounted GelSight sensors using an auto-encoder and concatenating the resulting feature vectors. 
The auto-encoder is trained with simulated sensor data from a set of synthesized objects.
To estimate the pose of a grasped object, the features extracted from real sensor data are compared against a set of features extracted from simulated sensor data obtained by simulating the same object in a wide variety of different in-hand poses. 

Many of these methods rely on neural networks or machine learning approaches, which require extensive training and large datasets. In contrast, our work employs a registration algorithm that eliminates the need for training, allowing flexible adaption to new objects without additional effort. Further, while other approaches focus on the refinement or tracking of object positions and depend on an accurate initial pose estimate, our registration algorithm performs one-shot pose estimation without requiring prior information about the object's pose.
\section{Proposed Method}
\label{chap:method}

Our proposed method for robotic in-hand pose estimation combines visual and tactile sensor data to achieve accurate and robust results. We preprocess point clouds obtained by a depth camera and vision-based tactile sensors, apply different weights to the individual points based on their source, and fuse the point clouds into one. We use an augmented Iterative Closest Point (ICP) algorithm, which we adapted to handle weighted point clouds, to obtain the pose estimation.
Fig.~\ref{fig:pipeline} provides an overview of our proposed method.

\begin{figure}
     \centering

     \includegraphics[width=0.5\linewidth]{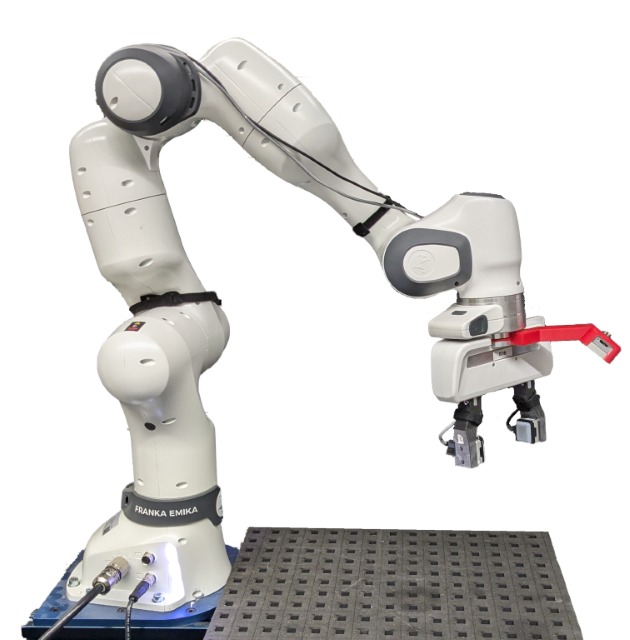}
     \hfill
     \includegraphics[width=0.45\linewidth]{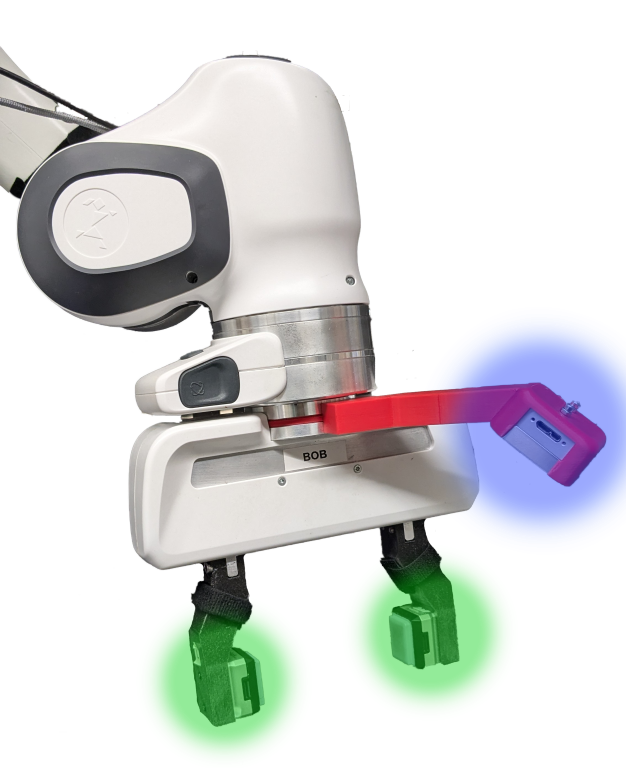}
     
     {\small\hspace{0.1 \textwidth}(a)\hspace{.45\textwidth}(b)}
     
     \caption{Robotic platform and sensors. We use the Franka Research 3 robotic platform to perform our experiments \textbf{(a)}. The working surface allows for objects to be fixed to the table, so they cannot move during pose estimation experiments. Figure \textbf{(b)} shows the Franka Hand parallel gripper with finger attachments for the GelSight Mini sensors (highlighted in green) and a wrist mount for the D405 depth camera (highlighted in blue).}
     \label{fig:setup}
\end{figure}

\begin{figure}
    \centering
    \includegraphics[width=\linewidth]{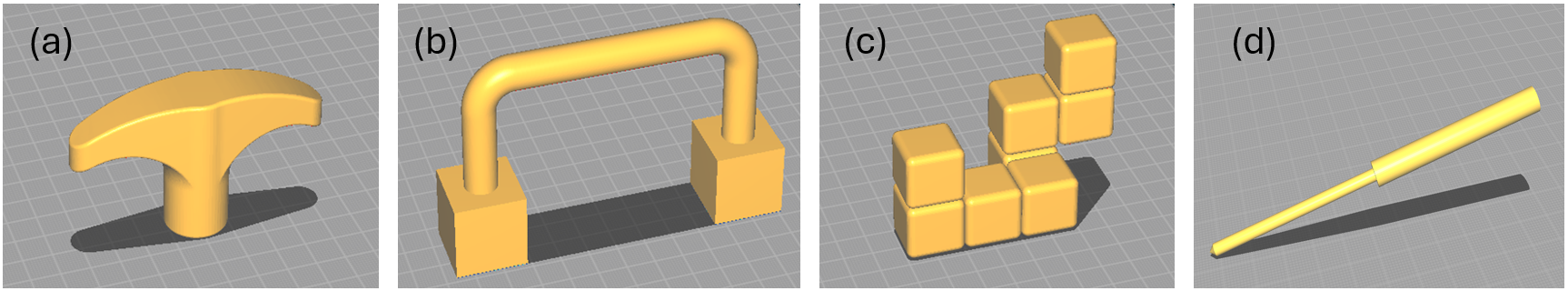}
    \caption{The objects used in our experiments \textbf{(a)} \textit{Knob}, \textbf{(b)} \textit{Handle}, \textbf{(c)} \textit{SL-block}, and \textbf{(d)} the \textit{Screwdriver} used for the insertion task. }
    \label{fig:objects}
\end{figure}

\subsection{Hardware Setup}

We use the Franka Research 3 robotic platform equipped with a parallel gripper in combination with an Intel RealSense D405 depth camera and two GelSight Mini vision-based tactile sensors (see Fig.~\ref{fig:setup}).
The camera is mounted to the wrist of the robotic arm facing toward the center of the gripper, while the tactile sensors are attached to the fingertips of the end effector. The camera and tactile sensor attachments are 3D-printed based on their publicly available 3D models\footnote{Realsense D405 camera mount:\\ \url{https://functional-manipulation-benchmark.github.io/files/index.html}}\footnote{GelSight finger attachment:\\ \url{https://www.gelsight.com/gelsightmini/}}.
Mounting the camera to the robotic wrist offers certain advantages over a stationary camera, by being in close proximity to the grasped object. Thus, the camera can capture the object at a higher resolution, and occlusion by other objects in the scene or by the robot itself can be reduced.

\subsection{Data Processing}

Both point clouds obtained from the camera and from the tactile sensors are downsampled to a density of $1~\text{point}/\text{mm}^3$ to ensure a consistent density for point clouds from different modalities during the pose estimation process. 
The depth and color stream of the camera is converted into a point cloud which is segmented and filtered in subsequent steps to obtain the points belonging to the object.
For our experiments, the object segmentation is limited to a color segmentation discarding all non-green points of the point cloud. To recognize differently colored objects, the integration of a more sophisticated segmentation method (for example \cite{wong2017segicp, ravi2024sam, wen2020robust}) would be necessary. 
Since the point cloud obtained by the depth camera can be noisy, we filter individual outlying points and average the point cloud over multiple successive frames from the camera.

Both fingers of the gripper are equipped with a GelSight Mini sensor, which provides details about the contact and local geometry of the grasped object. We reconstruct the deformation of the sensors' gel surface using the GelSight SDK\footnote{\url{https://github.com/gelsightinc/gsrobotics}} and generate a point cloud representing the local geometry of the grasped object.

\subsection{Sensor Fusion and Point Cloud Weighting}
\label{sec:sensor_fusion}

We obtain two point clouds, one from the tactile module, which provides a high-resolution and precise map of the contact surfaces between fingers and the object, and one from the camera module, which may cover more of the object's surface area but is also considerably noisier. 

Based on the sensor modality we assign a label to each point to track whether it is obtained from a tactile sensor or the depth camera. We also assign a weight to each individual point based on the point's modality to determine the influence each sensor modality should have during the subsequent pose estimation process. This weight can either be adjusted dynamically during the runtime or set to a suitable value in advance.

Once the weights are determined the sensor data is fused by concatenating the individual point clouds into a new single point cloud while keeping track of each point's corresponding sensor modality and the assigned weight.

We explore the possibility of automatically determining the weights based on certain characteristics of the point clouds. The underlying intuition is that under certain conditions, such as heavy occlusion or high noise levels in the vision point cloud,  tactile sensor data might be more reliable and should be weighted higher. For this purpose, we extract the following metrics from both the camera point cloud and the tactile point cloud. In our first experiment, we investigate whether and to what extent these metrics are suitable for determining an appropriate adaptive weighting.

\begin{itemize}

\item\textbf{Occlusion:}
The occlusion level quantifies the percentage of object surface area not captured by the vision module, including both physically obscured regions and areas missed due to environmental factors like challenging lighting conditions. 

\item\textbf{Noise:}
This metric quantifies the fluctuation in point clouds across consecutive measurements. Even when sensors and the object remain stationary, the obtained point clouds demonstrate certain temporal variations. 

\item\textbf{Number of points:}
The number of points in the visual and tactile point clouds serves as an indicator of the amount of information captured by each modality. 

\item\textbf{Point cloud volume:}
The point cloud volume quantifies the dispersion and spatial coverage of the visual and tactile point clouds. 

\end{itemize}

\subsection{Point Cloud Registration}

We estimate the object pose by aligning the mesh of the original object with the point cloud obtained from the employed sensors using an augmented Iterative Closest Point (ICP) algorithm. To integrate data from multiple sensor modalities, we assign weights to individual points, thereby adjusting each sensor’s contribution.

ICP, initially introduced by Chen and Medoni~\citet{chen1992icp} and Besl and McKay~\citet{besl1992icp}, iteratively aligns two point clouds by minimizing the distance between corresponding points. Its main steps are:
\begin{enumerate}
    \item \textbf{Initialization:} Initialize the target and source point clouds. In our case a point cloud sampled from the original mesh and the point cloud obtained from the sensors.
    \item \textbf{Correspondence Matching:} Identify the closest target point for each sensor point.
    \item \textbf{Transformation Estimation:} Compute and apply the transformation minimizing point-to-point distances.
    \item \textbf{Iteration:} Repeat until convergence or a set iteration limit.
\end{enumerate}

To handle fused point clouds from different sensors, we modify the transformation estimation by incorporating weights. Let $\mathbf{A}=\{a_1, \ldots, a_N\}$ and $\mathbf{B}=\{b_1, \ldots, b_N\}$ be the sets of corresponding points from the source and target point clouds respectively, with weights $\mathbf{w}=\{w_1, \ldots, w_N\}$. We first normalize the weights, then compute the weighted centroids $\bar{a}$ and $\bar{b}$, to align both sets by their centroids. 

To determine the optimal rotation that aligns both sets, we compute the weighted covariance matrix, defined as
$$
\mathbf{H} = \sum_{i=1}^{N} w_i (a_i-\bar{a})(b_i-\bar{b})^T,
$$
and perform singular value decomposition (SVD) on $\mathbf{H}$:
$$
\mathbf{H} = \mathbf{USV}^T, \quad \mathbf{R} = \mathbf{V}\mathbf{U}^T.
$$
The translation to realign the centroids is given by:
$$
\mathbf{t} = \bar{b} - \mathbf{R}\bar{a}.
$$
The final $4\times4$ transformation matrix is:
$$
\mathbf{T} = \begin{bmatrix} \mathbf{R} & \mathbf{t} \\ \mathbf{0} & 1 \end{bmatrix}.
$$
This weighted transformation is applied iteratively as part of the ICP process.

The success of the ICP algorithm is often determined by the quality of the initial transformation. Therefore, we perform multiple runs, rotating the initialized target point cloud in 90-degree increments around each principal axis, selecting the transformation with the smallest correspondence error.

\section{Experiments}
\label{chap:experiments}

\begin{figure}
    \centering
    \includegraphics[width=\linewidth]{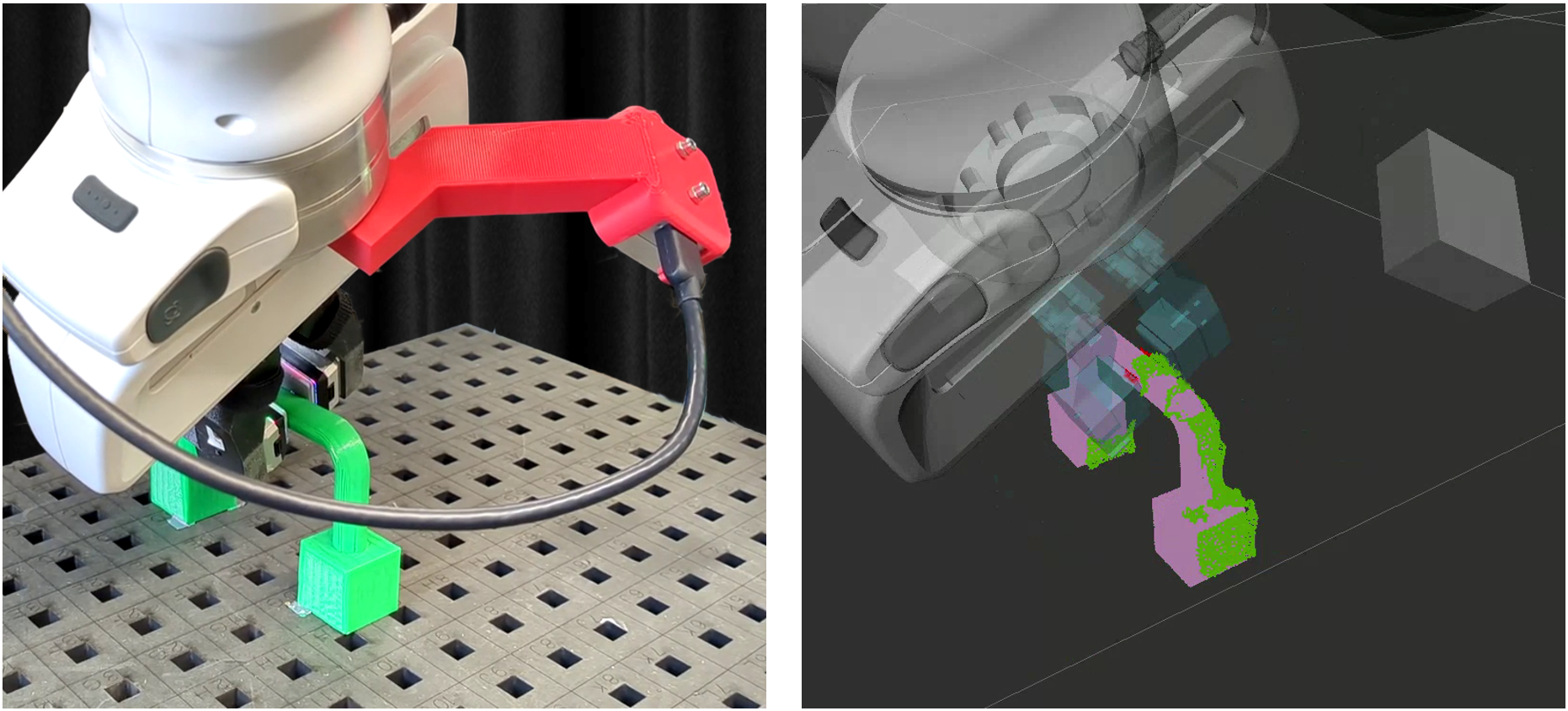}
     {\small\hspace{0.1 \textwidth}(a)\hspace{.45\textwidth}(b)}

    \caption{\textbf{(a)}: The robot grasping the object. \textbf{(b)}: Purple: Ground truth position of the object; Green: Point cloud obtained from RGB-D camera; Red: Point cloud obtained from the tactile sensors.}
    \label{fig:grasp}
    \small
\end{figure}

\begin{table*}[ht]
    \centering
    \caption{Success rate and average translation, rotation, and object error (Hausdorff distance) for each of the used objects. We compare the performance of our method using visual information only (\textbf{Vis}) with the performance when employing both the visual and tactile modality (\textbf{Vis + Tac}). The results on the handle object also include the results reported by Gao et al.~\cite{gao2023hand}. The rightmost columns contain the average errors of \textbf{successful} attempts only.}
    \label{tab:exp1b_success_accuracy}

    \renewcommand{\arraystretch}{0.7}
    \small
    \begin{tabular}{lrrrrrrrrr}\toprule
        &  &  & \multicolumn{3}{c}{Average Error} &  & \multicolumn{3}{c}{\parbox[t]{2.5cm}{\centering Average Error\\ \vspace{-0.15cm}{\tiny (successful~only)}}}\\
        \cmidrule{4-6} \cmidrule{8-10}
        & Success Rate &  & {\parbox[t]{1.1cm}{\tiny\centering Object Error [mm]}} & {\parbox[t]{1.1cm}{\tiny\centering Translation Error [mm]}} & {\parbox[t]{1.1cm}{\tiny\centering Rotation Error [deg]}} &  & {\parbox[t]{1.1cm}{\tiny\centering Object Error [mm]}} & {\parbox[t]{1.1cm}{\tiny\centering Translation Error [mm]}} & {\parbox[t]{1.1cm}{\tiny\centering Rotation Error [deg]}} \\
        \midrule
        \textbf{All Objects}&               &  &        &       &       &  &      &      &      \\
        Vis        & 117/159~~(74\%) &  &  15.08 &  9.44 & 21.09 &  & 5.14 & 3.28 & 3.90 \\
        Vis + Tac  & 127/159~~(\textbf{80\%}) &  &  \textbf{11.63} &  \textbf{7.50} & \textbf{16.70} &  & \textbf{4.54} & \textbf{2.87} & \textbf{3.53} \\
        \textbf{Knob}&              &  &        &       &       &  &      &      &      \\
        Vis        & 39/47~~(83\%)   &  &   8.67 &  6.48 & 11.64 &  & 5.05 & 4.03 & 5.18 \\
        Vis + Tac  & 40/47~~(\textbf{85\%})   &  &   \textbf{7.31} &  \textbf{5.77} & \textbf{10.08} &  & \textbf{4.13} & \textbf{3.15} & \textbf{4.58} \\
        \textbf{SL-Block}&          &  &        &       &       &  &      &      &      \\
        Vis        & 30/45~~(67\%)   &  &  13.61 &  8.92 & 31.42 &  & 3.31 & 1.53 & 3.31 \\
        Vis + Tac  & 33/45~~(\textbf{73\%})   &  &  \textbf{11.43} &  \textbf{7.55} & \textbf{27.38} &  & \textbf{3.19} & \textbf{1.47} & \textbf{3.24} \\
        \textbf{Handle}&            &  &        &       &       &  &      &      &      \\
        Vis        & 48/67~~(72\%)   &  &  20.38 & 11.76 & 20.50 &  & 6.37 & 3.77 & 3.23 \\
        Vis + Tac  & 54/67~~(\textbf{81\%})   &  &  \textbf{14.79} &  \textbf{8.68} & 14.18 &  & \textbf{5.68} & \textbf{3.51} & \textbf{2.94} \\
        Gao et al. & 32/45~~(71\%)   &  &    --- & 13.10 & \textbf{11.77} &  &  --- & 7.94 & 8.44 \\
        
        \bottomrule
    \end{tabular}
\end{table*}

We conducted several experiments to evaluate our in-hand object pose estimation method using multi-modal sensor data. The experiments address: (\textit{A}) optimal weighting of visual and tactile modalities, (\textit{B}) overall accuracy and reliability, including the effect of visual occlusion, and (\textit{C}) a practical insertion task. The objects used can be seen in Fig.~\ref{fig:objects}.

During the first two experiments the object is fixed on a work surface in a known position. The object's position and orientation can be adjusted to allow a broad range of gripping configurations. The robot grasps the object from different angles and positions (see Fig.~\ref{fig:grasp}). While the robot and object remain stationary, pose estimation is performed and the estimated pose is compared against the ground truth. We report the pose estimation error in terms of translation error, rotation error, and object error. We define the object error using the Hausdorff distance, measuring the maximum distance from any point on the ground truth mesh to the nearest point on the estimated pose mesh, and vice versa.
A vision-only baseline is obtained by repeating the estimation using vision-data only.

\subsection{Weighting Strategy}
\label{sec:exp1a}

In our first experiment, we investigate the optimal weighting ratio between visual and tactile sensor modalities. We vary the weight ratio between the visual (fixed at 1) and tactile (varied from 0.5 to 50) point clouds to determine the optimal combination for pose estimation.
We investigate whether the selected object influences the ideal weight ratio and whether point cloud metrics such as occlusion, noise levels, or point cloud volume can be leveraged to determine suitable weights dynamically.

\subsubsection{Object Specific Weighting}

\begin{figure}
    \centering
    \includegraphics[width=\linewidth]{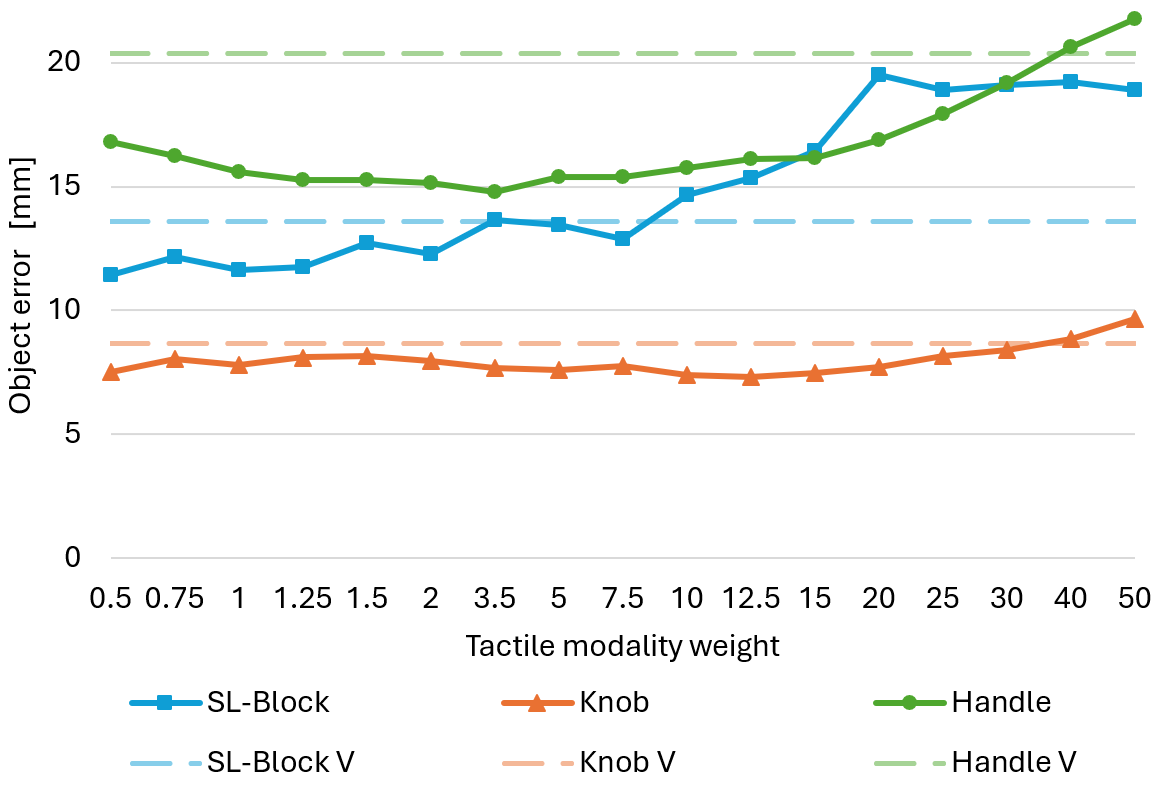}
    \caption{Average pose estimation error by weight. To determine appropriate weight ratios we varied the weight of the tactile point cloud while the weight of the vision point cloud remained 1. The solid lines show the average pose estimation error at each tested weight ratio. The dashed lines show the performance using only visual data.}
    \label{fig:exp1a_weights_error}
\end{figure}

Fig.~\ref{fig:exp1a_weights_error} shows that the selected weight significantly impacts pose estimation for the \textit{Handle} and \textit{SL-block} objects, while the \textit{Knob} object is less sensitive to the selected weights. For each object, several weight configurations outperform the vision-only baseline, though some result in higher errors.

\subsubsection{Dynamic Weighting}

We examined the correlation between optimal weights and the metrics \textit{occlusion, noise, number of points} and \textit{point cloud volume} for both the visual and the tactile point cloud. The metrics were extracted during each trial and the corresponding best weight ratio was determined based on the weights that resulted in the smallest pose estimation error for each particular grasping configuration. A weak positive correlation between the volume of the tactile point cloud and the optimal weight for the handle object ($r(65)$=$.27$, $p $=$.027$) exists, and a moderate negative correlation between visual point cloud volume and the optimal weight for the SL-block ($r(43)$=$.34$, $p$=$.024$). Other metrics did not show significant correlations.

\begin{figure*}
\BottomFloatBoxes
    \begin{floatrow}
    \ffigbox[0.3\textwidth]{%
        \includegraphics[width=0.75\linewidth]{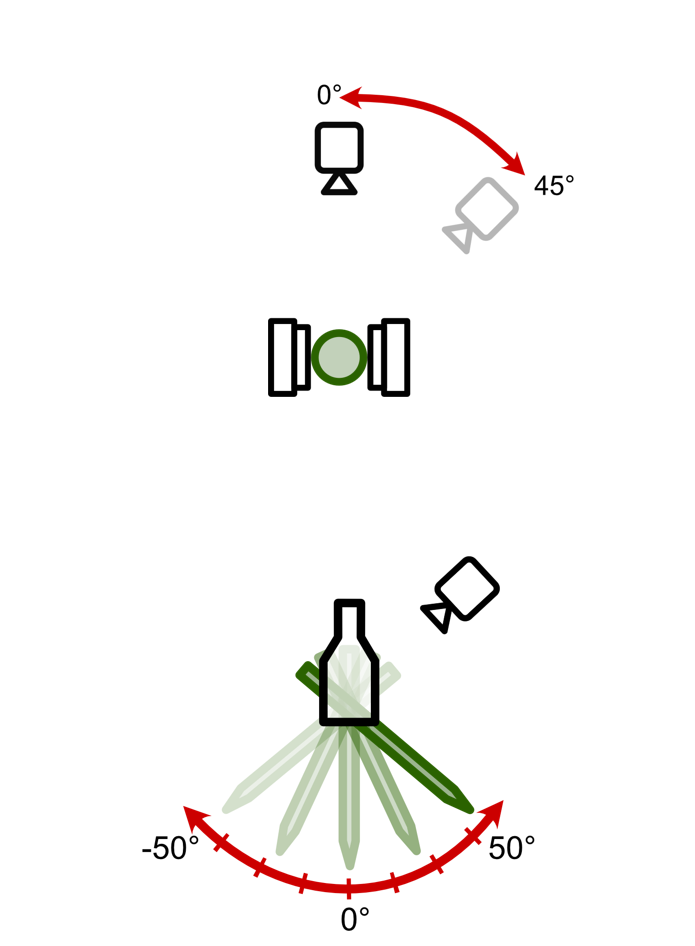}
        \put(-120,147){\footnotesize (a) Camera variations (top view):}
        \put(-120, 73){\footnotesize (b) Tool angle variations (side view):}
    }{%
        \caption{Variations of the insertion task experiment: \textbf{(a)} The experiment is conducted with the camera pointing either straight between the fingers (0° around the z-axis) or at an angle (45° around the z-axis). \textbf{(b)} The screwdriver's angle between the fingers is varied from -50° to +50° in 5° increments.%
        \label{fig:insertion_variations}}%
    }

    \hfill
    \capbtabbox[0.63\textwidth]{%
    \small
    \renewcommand{\arraystretch}{0.7}
    {\begin{tabular}{@{}rrlcrlrcrlcrlr@{}}
    \toprule
         &  \multicolumn{6}{c}{Camera at 0°} & \phantom{123} & \multicolumn{6}{c}{Camera at 45°} \\
         \cmidrule{2-7} \cmidrule{9-14}
         Tool Angle \phantom{1 } & \multicolumn{2}{c}{~~Vis~~}     &  & \multicolumn{2}{c}{Vis+Tac} & Occ.       &  & \multicolumn{2}{c}{~~Vis~~}     &  & \multicolumn{2}{c}{Vis+Tac}     & Occ. \\
         \midrule
           0° \phantom{12345} & \checkmark      & \checkmark      &  & \checkmark      & \checkmark      & 80\% &  & \checkmark     & \checkmark     &  & \checkmark     & \checkmark     & 96\% \\
          -5° \phantom{12345} & \checkmark      & \checkmark      &  & \checkmark      & \checkmark      & 82\% &  & \checkmark     & \checkmark     &  & \checkmark     & \checkmark     & 95\% \\
         -10° \phantom{12345} & \checkmark      & \checkmark      &  & \checkmark      & \checkmark      & 85\% &  & \checkmark     & \texttimes     &  & \checkmark     & \checkmark     & 94\% \\
         -15° \phantom{12345} & \checkmark      & \checkmark      &  & \textopenbullet & \checkmark      & 86\% &  & \texttimes     & \texttimes     &  & \texttimes     & \checkmark     & 94\% \\
         -20° \phantom{12345} & \textopenbullet & \textopenbullet &  & \checkmark      & \checkmark      & 86\% &  & \texttimes     & \texttimes     &  & \checkmark     & \checkmark     & 94\% \\
         -25° \phantom{12345} & \textopenbullet & \checkmark      &  & \checkmark      & \checkmark      & 88\% &  & \texttimes     & \texttimes     &  & \checkmark     & \checkmark     & 94\% \\
         -30° \phantom{12345} & \textopenbullet & \textopenbullet &  & \checkmark      & \textopenbullet & 89\% &  & \texttimes     & \texttimes     &  & \checkmark     & \checkmark     & 94\% \\
         -35° \phantom{12345} & \textopenbullet & \texttimes      &  & \checkmark      & \textopenbullet & 93\% &  & \texttimes     & \texttimes     &  & \checkmark     & \checkmark     & 94\% \\
         -40° \phantom{12345} & \texttimes      & \texttimes      &  & \texttimes      & \checkmark      & 98\% &  & \texttimes     & \texttimes     &  & \checkmark     & \checkmark     & 94\% \\
         -45° \phantom{12345} & \textopenbullet & \texttimes      &  & \checkmark      & \checkmark      & 98\% &  & \texttimes     & \texttimes     &  & \checkmark     & \checkmark     & 94\% \\
         -50° \phantom{12345} & \texttimes      & \texttimes      &  & \checkmark      & \checkmark      & 98\% &  & \texttimes     & \texttimes     &  & \textopenbullet & \checkmark     & 94\% \\
    \bottomrule
    \end{tabular}}
    \includegraphics[width=0.8\linewidth]{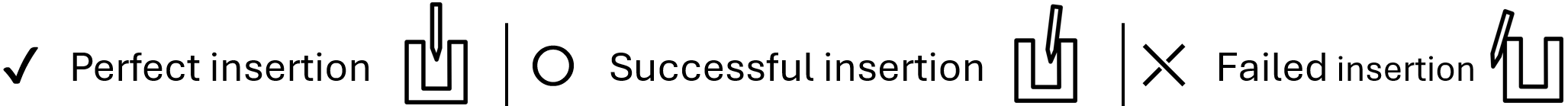}
    }{%
      \caption{Results from the insertion task experiment for varying tool angles under the two camera configurations (0$^\circ$ and 45$^\circ$ z-rotation). The table compares the performance of using vision data only (Vis) and the combination of visual and tactile information (Vis+Tac) for our pose estimation method. For each condition, two insertion attempts were made. 
      \label{tab:insertion_results}}%
    }
    \end{floatrow} 
\end{figure*}

\subsection{Object Pose Estimation}
\label{sec:exp1b}

Our second experiment investigates the accuracy and reliability of our method and compares its performance with that of the approach proposed by Gao et al.~\cite{gao2023hand}. In addition, we investigate the impact of tactile sensors, particularly in the context of visual object occlusion. 
We use the following vision-to-tactile ratios determined in the previous experiment \textbf{Knob:} 1:12.5, \textbf{Handle:} 1:3.5, \textbf{SL-Block:} 1:0.5. A pose estimation is considered \textit{successful} if the translation error is below 15~mm and the rotation error below 15~degrees.

\subsubsection{Success Rate and Accuracy}

\begin{figure}
    \centering
    \includegraphics[width=\linewidth]{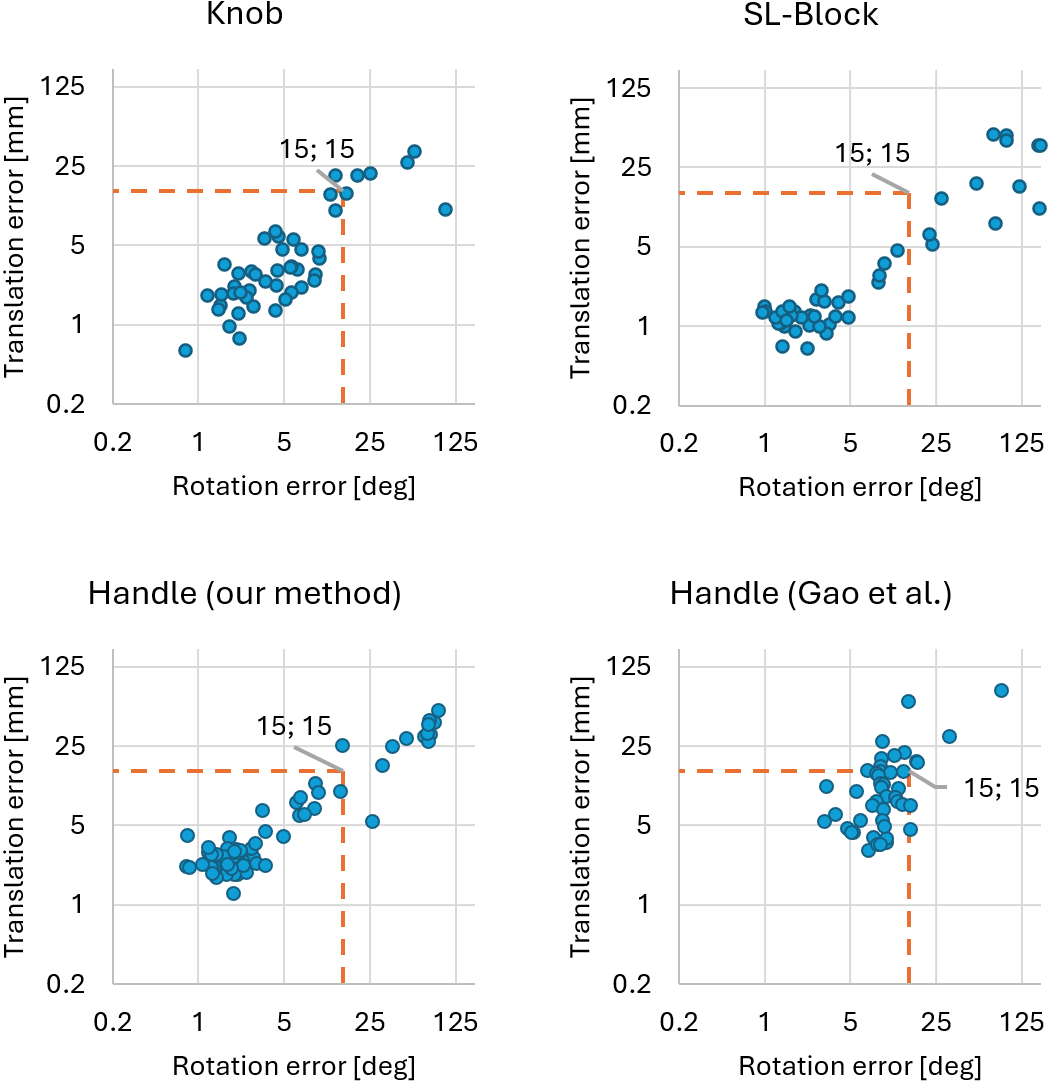}
    \caption{Rotation and translation error of each pose estimation attempt on the knob, handle, and SL-block using the object-specific weight. Pose estimations within the dashed lines are considered to be \textit{successful}. The bottom-right plot shows the result Gao et al. \cite{gao2023hand} report using their method.}
    \label{fig:exp1b_rot_trans_error}
\end{figure}

Fig.~\ref{fig:exp1b_rot_trans_error} shows that while errors range from sub-millimeter accuracy up to a 50~mm translation error and nearly 180 degree rotation error respectively, most remain below 5~mm and 5 degrees. As summarized in Table~\ref{tab:exp1b_success_accuracy}, by adding tactile data the overall success rate increases from 74\% (visual information only), to 80\%.  Moreover, the supplementary tactile information reduces errors for all objects. The handle object shows the greatest benefit from the additional tactile modality. The average error for \textit{successful} pose estimations only, shows minimal improvement for the SL-block and handle object, while the translation error for the knob object improved by around 1~mm with additional tactile sensing.

\subsubsection{Influence of Occlusion}

\begin{figure}
    \centering
    \includegraphics[width=\linewidth]{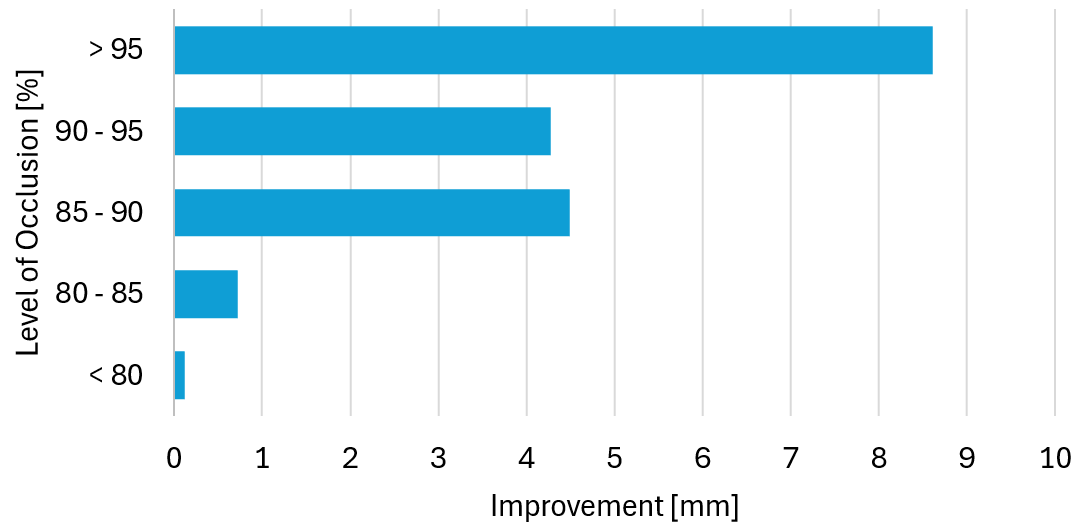}
    \caption{The average improvement achieved by combining tactile and visual information compared to using visual information alone.}
    \label{fig:exp1b_error_improvements}
\end{figure}

Using vision alone, there is a strong positive correlation between object error and occlusion ($r(157)=.619$, $p<.001$).
With fused visual and tactile information, this relationship is similarly strong, indicating that occlusion remains a major factor despite the addition of the tactile modality.
Although the error generally increases with occlusion, Fig. \ref{fig:exp1b_error_improvements} shows that the benefit of adding tactile sensing also increases.

\subsubsection{Comparison with Gao et al.~\cite{gao2023hand}}

We included the handle object used in the work of Gao et al.~\cite{gao2023hand} in our experiments to allow a comparison between their and our method (see Table~\ref{tab:exp1b_success_accuracy}).
Using our method, we could achieve a higher success rate and looking at successful attempts only, our method produces on average significantly lower translation and rotation errors.
Across all trials, our method provides a lower translation error but performs worse with respect to the rotation error.
The individual errors displayed in Fig.~\ref{fig:exp1b_rot_trans_error} show that our method exhibits several failed attempts with errors beyond 50~mm and 50 degrees, while most errors in Gao et al.'s results remain below 20~mm or 20~degrees.

\subsection{Insertion Task}
\label{sec:exp2}

\begin{figure}
    \centering
    \includegraphics[width=\linewidth]{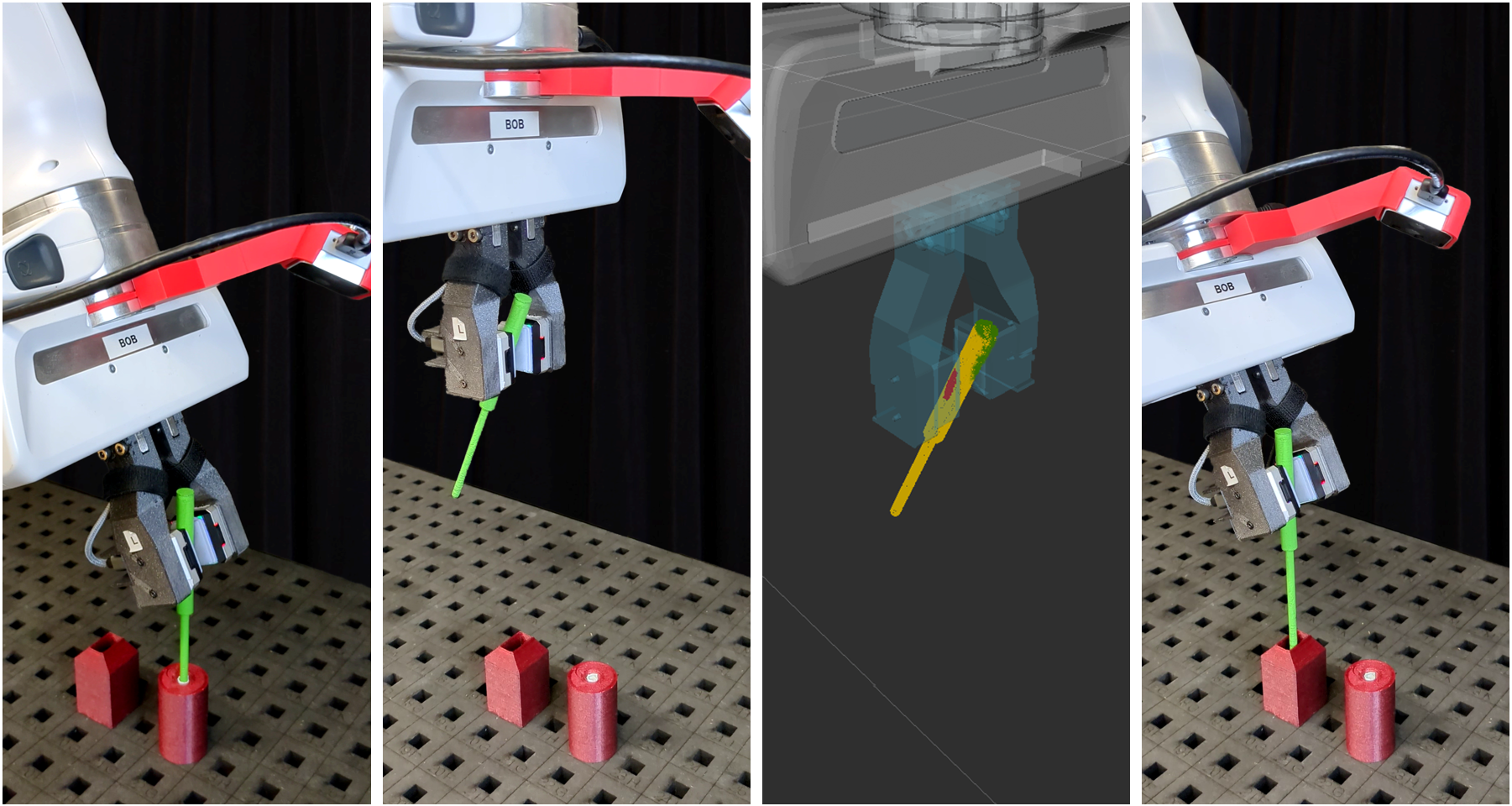}
    \small
    ~(a)\hfill(b)\hfill(c)\hfill(d)\hfill \phantom{~}
    \caption{Sequence of the insertion task: (a) The screwdriver is located in a known position and the robot picks it up at a predetermined angle (in this case -25 degrees). (b) Without prior knowledge about the pickup angle and location, pose estimation of the tool is performed. Figure (c) shows the visualization of the point clouds during pose estimation: Red: tactile point cloud, Green: vision point cloud, Yellow: estimated tool pose. (d) After the pose is estimated the robot aligns the tool with the hole and attempts insertion.}
    \label{fig:insertion}
\end{figure}

In our final experiment we want to verify our method in a practical real-world task and show the ability of tactile data to compensate for the lack of visual data in cases of high occlusion. The robot grasps a screwdriver, estimates its pose, and attempts to insert it into a 1~cm hole (see Fig.~\ref{fig:insertion}). Insertion quality is classified as \textit{perfect} (no contact with the hole perimeter), \textit{successful} (edge contact but insertion achieved), or \textit{failed}. By varying the screwdriver's angle within the gripper, we control the degree of visual occlusion and by changing the position of the camera, we can vary the amount of occlusion caused by the robot's fingers. 
There are two camera conditions as shown in Fig. \ref{fig:insertion_variations}.

In cases where the screwdriver is angled towards the camera (positive tool angles), resulting in low occlusion levels, our method achieves perfect insertion in both vision-only and visuotactile conditions. Therefore we will focus on the insertion results of tool angles varying from 0° to -50° (see \autoref{tab:insertion_results}).

\subsubsection{Camera at 0° Position}
With the camera in the 0-degree position, the results show a clear trend where decreasing the angle of the screwdriver past 0 degrees leads to higher levels of occlusion. Without the additional information from the tactile sensors, this also leads to a decrease in insertion success. However, by using data obtained from both vision and touch, our method can maintain a high success rate throughout the varying angles. 

\subsubsection{Camera at 45° Position}
The altered viewing angle at 45 degrees around the z-axis leads to high occlusion levels ($94-96\%$) across all trials, as the fingers of the gripper block large parts of the camera's view on the tool. Vision-only insertion attempts were mostly unsuccessful with negative angles of the tool. Only at small angles between 0 degrees and -10 degrees \textit{perfect} insertions were observed. At all other negative angles, the pose estimation was not accurate enough, leading to \textit{failed} insertion attempts.

By adding the tactile modality the performance was again greatly improved, providing \textit{perfect} insertions across all angles, with only a few exceptions of \textit{successful} and \textit{failed} attempts.

\section{Discussion}
\label{chap:discussion}

Our experiments validate the effectiveness of integrating visual and tactile sensing for in-hand pose estimation, particularly under high occlusion. While vision-based methods suffice when occlusion is low, the addition of tactile data significantly enhances accuracy in challenging conditions.

The first experiment demonstrated that balancing the contribution of visual and tactile information is crucial. Object-specific properties such as shape and geometric features may influence the optimal weight ratios. The handle and SL-block benefit more from tactile data than the knob, possibly due to their geometric properties. The SL-block, for instance, has several repeating visual features that might require the additional tactile input to resolve pose ambiguities. Dynamic weighting strategies showed potential, but no universally applicable model was identified, emphasizing the need for further research in adaptive sensor fusion.

Our method improved success rates from 74\% to 80\% by incorporating tactile data. While most errors remained below 5~mm and 5~degrees, the presence of outliers with large rotation and translation errors suggests that refining initialization strategies could further improve performance and robustness. 

The comparison with results from Gao et al.~\cite{gao2023hand} showed that our method is able to achieve a similar pose estimation accuracy, without relying on machine learning techniques or the collection of large data sets. Our method even outperforms the compared model in terms of the success rate.

Occlusion strongly correlated with pose estimation error. Although incorporating tactile data mitigated some of these errors, occlusion remained a significant factor. Our results indicate that, while higher occlusion generally increases error, the benefits of tactile data also grow. Some configurations with high occlusion still yielded accurate estimations, highlighting the importance of distinctive object features rather than just visibility.

\bibliographystyle{IEEEtran}
\bibliography{lit}

\begin{thebibliography}{10}
\providecommand{\url}[1]{#1}
\csname url@samestyle\endcsname
\providecommand{\newblock}{\relax}
\providecommand{\bibinfo}[2]{#2}
\providecommand{\BIBentrySTDinterwordspacing}{\spaceskip=0pt\relax}
\providecommand{\BIBentryALTinterwordstretchfactor}{4}
\providecommand{\BIBentryALTinterwordspacing}{\spaceskip=\fontdimen2\font plus
\BIBentryALTinterwordstretchfactor\fontdimen3\font minus
  \fontdimen4\font\relax}
\providecommand{\BIBforeignlanguage}[2]{{%
\expandafter\ifx\csname l@#1\endcsname\relax
\typeout{** WARNING: IEEEtran.bst: No hyphenation pattern has been}%
\typeout{** loaded for the language `#1'. Using the pattern for}%
\typeout{** the default language instead.}%
\else
\language=\csname l@#1\endcsname
\fi
#2}}
\providecommand{\BIBdecl}{\relax}
\BIBdecl

\bibitem{litvak2019learning}
Y.~Litvak, A.~Biess, and A.~Bar-Hillel, ``Learning pose estimation for
  high-precision robotic assembly using simulated depth images,'' in \emph{2019
  International Conference on Robotics and Automation (ICRA)}.\hskip 1em plus
  0.5em minus 0.4em\relax IEEE, 2019, pp. 3521--3527.

\bibitem{quentin2023industrial}
P.~Quentin, D.~Knoll, and D.~Goehring, ``Industrial application of 6d pose
  estimation for robotic manipulation in automotive internal logistics,'' in
  \emph{2023 IEEE 19th International Conference on Automation Science and
  Engineering (CASE)}.\hskip 1em plus 0.5em minus 0.4em\relax IEEE, 2023, pp.
  1--8.

\bibitem{billard2019trends}
A.~Billard and D.~Kragic, ``Trends and challenges in robot manipulation,''
  \emph{Science}, vol. 364, no. 6446, p. eaat8414, 2019.

\bibitem{wen2020robust}
B.~Wen, C.~Mitash, S.~Soorian, A.~Kimmel, A.~Sintov, and K.~E. Bekris,
  ``Robust, occlusion-aware pose estimation for objects grasped by adaptive
  hands,'' in \emph{2020 IEEE International Conference on Robotics and
  Automation (ICRA)}.\hskip 1em plus 0.5em minus 0.4em\relax IEEE, 2020, pp.
  6210--6217.

\bibitem{izatt2017tracking}
G.~Izatt, G.~Mirano, E.~Adelson, and R.~Tedrake, ``Tracking objects with point
  clouds from vision and touch,'' in \emph{2017 IEEE International Conference
  on Robotics and Automation (ICRA)}.\hskip 1em plus 0.5em minus 0.4em\relax
  IEEE, 2017, pp. 4000--4007.

\bibitem{liu2023enhancing}
Y.~Liu, X.~Xu, W.~Chen, H.~Yuan, H.~Wang, J.~Xu, R.~Chen, and L.~Yi,
  ``Enhancing generalizable 6d pose tracking of an in-hand object with tactile
  sensing,'' \emph{IEEE Robotics and Automation Letters}, 2023.

\bibitem{anzai2020deep}
T.~Anzai and K.~Takahashi, ``Deep gated multi-modal learning: In-hand object
  pose changes estimation using tactile and image data,'' in \emph{2020
  IEEE/RSJ International Conference on Intelligent Robots and Systems
  (IROS)}.\hskip 1em plus 0.5em minus 0.4em\relax IEEE, 2020, pp. 9361--9368.

\bibitem{dikhale2022visuotactile}
S.~Dikhale, K.~Patel, D.~Dhingra, I.~Naramura, A.~Hayashi, S.~Iba, and
  N.~Jamali, ``Visuotactile 6d pose estimation of an in-hand object using
  vision and tactile sensor data,'' \emph{IEEE Robotics and Automation
  Letters}, vol.~7, no.~2, pp. 2148--2155, 2022.

\bibitem{gao2023hand}
Y.~Gao, S.~Matsuoka, W.~Wan, T.~Kiyokawa, K.~Koyama, and K.~Harada, ``In-hand
  pose estimation using hand-mounted rgb cameras and visuotactile sensors,''
  \emph{IEEE Access}, vol.~11, pp. 17\,218--17\,232, 2023.

\bibitem{yuan2017gelsight}
W.~Yuan, S.~Dong, and E.~H. Adelson, ``Gelsight: High-resolution robot tactile
  sensors for estimating geometry and force,'' \emph{Sensors}, vol.~17, no.~12,
  p. 2762, 2017.

\bibitem{lambeta2020digit}
M.~Lambeta, P.-W. Chou, S.~Tian, B.~Yang, B.~Maloon, V.~R. Most, D.~Stroud,
  R.~Santos, A.~Byagowi, G.~Kammerer \emph{et~al.}, ``Digit: A novel design for
  a low-cost compact high-resolution tactile sensor with application to in-hand
  manipulation,'' \emph{IEEE Robotics and Automation Letters}, vol.~5, no.~3,
  pp. 3838--3845, 2020.

\bibitem{li2018deepim}
Y.~Li, G.~Wang, X.~Ji, Y.~Xiang, and D.~Fox, ``Deepim: Deep iterative matching
  for 6d pose estimation,'' in \emph{Proceedings of the European Conference on
  Computer Vision (ECCV)}, 2018, pp. 683--698.

\bibitem{zakharov2019dpod}
S.~Zakharov, I.~Shugurov, and S.~Ilic, ``Dpod: 6d pose object detector and
  refiner,'' in \emph{Proceedings of the IEEE/CVF international conference on
  computer vision}, 2019, pp. 1941--1950.

\bibitem{besl1992method}
P.~J. Besl and N.~D. McKay, ``Method for registration of 3-d shapes,'' in
  \emph{Sensor fusion IV: control paradigms and data structures}, vol.
  1611.\hskip 1em plus 0.5em minus 0.4em\relax Spie, 1992, pp. 586--606.

\bibitem{xiang2017posecnn}
Y.~Xiang, T.~Schmidt, V.~Narayanan, and D.~Fox, ``Posecnn: A convolutional
  neural network for 6d object pose estimation in cluttered scenes,''
  \emph{arXiv preprint arXiv:1711.00199}, 2017.

\bibitem{tekin2018real}
B.~Tekin, S.~N. Sinha, and P.~Fua, ``Real-time seamless single shot 6d object
  pose prediction,'' in \emph{Proceedings of the IEEE conference on computer
  vision and pattern recognition}, 2018, pp. 292--301.

\bibitem{raessa2019visually}
M.~Raessa, D.~Petit, W.~Wan, and K.~Harada, ``Visually guided extrinsic
  manipulation for assembly tasks,'' in \emph{2019 IEEE 4th International
  Conference on Advanced Robotics and Mechatronics (ICARM)}.\hskip 1em plus
  0.5em minus 0.4em\relax IEEE, 2019, pp. 202--207.

\bibitem{liu2021robust}
R.~Liu, W.~Wan, K.~Koyama, and K.~Harada, ``Robust robotic 3-d drawing using
  closed-loop planning and online picked pens,'' \emph{IEEE Transactions on
  Robotics}, vol.~38, no.~3, pp. 1773--1792, 2021.

\bibitem{pfanne2018fusing}
M.~Pfanne, M.~Chalon, F.~Stulp, and A.~Albu-Sch{\"a}ffer, ``Fusing joint
  measurements and visual features for in-hand object pose estimation,''
  \emph{IEEE Robotics and Automation Letters}, vol.~3, no.~4, pp. 3497--3504,
  2018.

\bibitem{galaiya2023exploring}
V.~R. Galaiya, M.~Asfour, T.~E. Alves~de Oliveira, X.~Jiang, and V.~Prado~da
  Fonseca, ``Exploring tactile temporal features for object pose estimation
  during robotic manipulation,'' \emph{Sensors}, vol.~23, no.~9, p. 4535, 2023.

\bibitem{driels1986pose}
M.~Driels, ``Pose estimation using tactile sensor data for assembly
  operations,'' in \emph{Proceedings. 1986 IEEE International Conference on
  Robotics and Automation}, vol.~3.\hskip 1em plus 0.5em minus 0.4em\relax
  IEEE, 1986, pp. 1255--1261.

\bibitem{lach2023placing}
L.~Lach, N.~Funk, R.~Haschke, S.~Lemaignan, H.~J. Ritter, J.~Peters, and
  G.~Chalvatzaki, ``Placing by touching: An empirical study on the importance
  of tactile sensing for precise object placing,'' in \emph{2023 IEEE/RSJ
  International Conference on Intelligent Robots and Systems (IROS)}.\hskip 1em
  plus 0.5em minus 0.4em\relax IEEE, 2023, pp. 8964--8971.

\bibitem{li2014localization}
R.~Li, R.~Platt, W.~Yuan, A.~Ten~Pas, N.~Roscup, M.~A. Srinivasan, and
  E.~Adelson, ``Localization and manipulation of small parts using gelsight
  tactile sensing,'' in \emph{2014 IEEE/RSJ International Conference on
  Intelligent Robots and Systems}.\hskip 1em plus 0.5em minus 0.4em\relax IEEE,
  2014, pp. 3988--3993.

\bibitem{bauza2023tac2pose}
M.~Bauza, A.~Bronars, and A.~Rodriguez, ``Tac2pose: Tactile object pose
  estimation from the first touch,'' \emph{The International Journal of
  Robotics Research}, vol.~42, no.~13, pp. 1185--1209, 2023.

\bibitem{caddeo2023collision}
G.~M. Caddeo, N.~A. Piga, F.~Bottarel, and L.~Natale, ``Collision-aware in-hand
  6d object pose estimation using multiple vision-based tactile sensors,'' in
  \emph{2023 IEEE International Conference on Robotics and Automation
  (ICRA)}.\hskip 1em plus 0.5em minus 0.4em\relax IEEE, 2023, pp. 719--725.

\bibitem{berk2015ycb}
B.~Calli, A.~Singh, A.~Walsman, S.~Srinivasa, P.~Abbeel, and A.~M. Dollar,
  ``The ycb object and model set: Towards common benchmarks for manipulation
  research,'' in \emph{2015 International Conference on Advanced Robotics
  (ICAR)}, 2015, pp. 510--517.

\bibitem{chaudhury2022using}
A.~N. Chaudhury, T.~Man, W.~Yuan, and C.~G. Atkeson, ``Using collocated vision
  and tactile sensors for visual servoing and localization,'' \emph{IEEE
  Robotics and Automation Letters}, vol.~7, no.~2, pp. 3427--3434, 2022.

\bibitem{wong2017segicp}
J.~M. Wong, V.~Kee, T.~Le, S.~Wagner, G.-L. Mariottini, A.~Schneider,
  L.~Hamilton, R.~Chipalkatty, M.~Hebert, D.~M. Johnson \emph{et~al.},
  ``Segicp: Integrated deep semantic segmentation and pose estimation,'' in
  \emph{2017 IEEE/RSJ International Conference on Intelligent Robots and
  Systems (IROS)}.\hskip 1em plus 0.5em minus 0.4em\relax IEEE, 2017, pp.
  5784--5789.

\bibitem{ravi2024sam}
N.~Ravi, V.~Gabeur, Y.-T. Hu, R.~Hu, C.~Ryali, T.~Ma, H.~Khedr, R.~R{\"a}dle,
  C.~Rolland, L.~Gustafson \emph{et~al.}, ``Sam 2: Segment anything in images
  and videos,'' \emph{arXiv preprint arXiv:2408.00714}, 2024.

\bibitem{chen1992icp}
Y.~Chen and G.~Medioni, ``Object modelling by registration of multiple range
  images,'' \emph{Image and vision computing}, vol.~10, no.~3, pp. 145--155,
  1992.

\bibitem{besl1992icp}
P.~J. Besl and N.~D. McKay, ``Method for registration of 3-d shapes,'' in
  \emph{Sensor fusion IV: control paradigms and data structures}, vol.
  1611.\hskip 1em plus 0.5em minus 0.4em\relax Spie, 1992, pp. 586--606.

\end{thebibliography}

\end{document}